\documentclass[preprint]{elsarticle}

\usepackage{graphicx}
\usepackage{multicol}
\usepackage{footmisc}

\usepackage{longtable}
\usepackage[export]{adjustbox}
\usepackage{subfig}
\usepackage{amsmath}
\usepackage{amssymb}
\usepackage{multirow}
\usepackage{multicol}
\usepackage[T2A]{fontenc}
\usepackage[utf8]{inputenc}
\usepackage[russian,english]{babel}
\usepackage{marvosym}
\usepackage{algorithm}
\usepackage{algpseudocode}

\usepackage{url}

\mathchardef\mhyphen="2D

\usepackage{lineno,hyperref}
\modulolinenumbers[5]

\journal{Cognitive Systems Research}

\biboptions{authoryear}

\makeatletter
\def\ps@pprintTitle{%
 \let\@oddhead\@empty
 \let\@evenhead\@empty
 \def\@oddfoot{}%
 \let\@evenfoot\@oddfoot}
\makeatother

\begin{document}
\let\today\relax
\begin{frontmatter}

\title{Complexity of Symbolic Representation in Working Memory of Transformer Correlates with the Complexity of a Task}

\author[mipt]{Alsu Sagirova\corref{correspondingauthor}}
\ead{alsu.sagirova@phystech.edu}

\author[mipt,airi]{Mikhail Burtsev}
\ead{burtcev.ms@mipt.ru}

\cortext[correspondingauthor]{Corresponding author}

\address[mipt]{Neural Networks and Deep Learning Lab,\\
Moscow Institute of Physics and Technology\\Institutskiy pereulok, 9, Dolgoprudny, 141701, Russia\\
}
\address[airi]{AIRI\\4B 08, Kutuzovsky prospect 32 build. 1, Moscow, 121170, Russia 
}

\begin{abstract}
Even though Transformers are extensively used for Natural Language Processing tasks, especially for machine translation, they lack an explicit memory to store key concepts of processed texts. This paper explores the properties of the content of symbolic working memory added to the Transformer model decoder. Such working memory enhances the quality of model predictions in machine translation task and works as a neural-symbolic representation of information that is important for the model to make correct translations. The study of memory content revealed that translated text keywords are stored in the working memory, pointing to the relevance of memory content to the processed text. Also, the diversity of tokens and parts of speech stored in memory correlates with the complexity of the corpora for machine translation task.
\end{abstract}


\end{frontmatter}


\section{Introduction}

Working memory is a theoretical concept from neuroscience that represents a cognitive system that temporarily stores information and manipulates it \citep{miyake:1999}. While neural network models successfully tackle various tasks in different fields of artificial intelligence and are robust to the data flaws, they still lack interpretability and generalization. Symbolic methods refer to explicit symbol data representation and processing. Therefore, symbolic methods are explainable and easy to check for correctness. Combining the neural network approach with symbolic components should empower neural model decision-making procedures and enhance the overall model performance.

Today neuro-symbolic architectures provide foundation for the area of Natural Language Processing (NLP). Here, the dominating approach is to train a model to encode a symbolic input into internal vector representations accumulated in some memory and then decode the content of this memory back to a sequence of symbols. The majority of NLP models have no explicit memory storage for the information missing in the input sequence. But for better processing of a text, additional contextual knowledge should be helpful. We hypothesize that operating with such knowledge associated with an input but not explicitly presented in it should help the model to understand the processed text conceptually and improve the quality of predictions. We propose to add symbolic working memory to the Transformer decoder to generate and store contextual knowledge in a textual form to simulate some sort of "inner speech."

\section{Related work}

The use of dedicated memory storage in neural network models to store and retrieve explicit or implicit vector representations required for computations is studied in the Memory-augmented neural networks (MANN) research area. The simplest examples of memory-augmented neural network models are Recurrent Neural Network (RNN) and Long-Short Term Memory (LSTM) \citep{Hochreiter:1997}, where the internal memory is represented by the model's hidden states, which summarize the input history and are controlled by the gates. 

External memory was implemented in the Neural Turing Machine (NTM) \citep{Graves:2014} and its successor, the Differentiable Neural Computer (DNC) \citep{Graves:2016}. The NTM has a memory matrix of a fixed size controlled by the neural network to store real-valued vector representations. The controller network receives inputs and produces outputs, and it also manipulates memory with parallel read and write heads. 

The DNC is an end-to-end differentiable extension of an NTM that uses the random-access memory concept. The controller network employs three attention mechanisms in read and write heads to interact with memory. The first mechanism is a content lookup, where the attention weights are based on a cosine similarity between the controller-generated key vector and each memory slot. The second mechanism is writing via dynamic memory allocation when the controller can free memory slots that are no longer required. The third mechanism is a temporal memory linkage used for sequential reading from memory slots. In the Sparse DNC model \citep{Rae:2016}, the authors test randomized k-d trees and locality sensitive hash (LSH) algorithms to make memory addressing sparse.

Memory Networks \citep{Weston:2014} and End-to-End Memory Networks \citep{Sukhbaatar:2015} employ a recurrent attention mechanism for reading memory to solve a question answering (QA) task. The input sequence embedding is stored in the memory and then matched with the query embedding to obtain attention scores. These scores are then applied to another representation of the input sequence to give a response vector. The model handles multi-hop memory updates by combining the previous layer input representation and the response vector into the current layer input representation. Hierarchical Memory Networks (HMNs) \citep{Chandar:2016} organize memory cells into a hierarchical structure to ease computation compared with the soft attention over flat memory. Also, unlike Memory Networks, which use soft attention entirely, HMNs apply soft attention for a subset of memory slots selected by a mechanism based on Maximum Inner Product Search (MIPS).

Dynamic NTMs (D-NTMs) \citep{Gulcehre:2017} extend an NTM model with a trainable scheme for memory addressing to allow various soft and hard attention mechanisms to read from memory. Reading is done in a multi-hop manner compared with the multi-head reading in NTMs. Also, feedforward and Gated Recurrent Unit (GRU) controller networks are tested.

The TARDIS \citep{Gulcehre:2017memory} memory structure is similar to NTMs and D-NTMs. For more efficient gradient propagation, memory is considered as storage for wormhole connections. Read and write operations are implemented with discrete addressing, and once memory is full, a heuristic is used for memory read and write operations.

The Global Context Layer (GCL) \citep{Meng:2018} incorporates the global context information into memory. The reading mechanism has separated address and content parts to ease the training. Compared to the NTM, GCL does not interpolate the attention vector at the current time step with the one from the previous time step but completely ignores it.

Transformer \citep{Vaswani:2017} is an encoder-decoder neural network model, which successfully solves various natural language processing tasks \citep{Raffel:2020}. The model is based on the attention mechanism that uses the information about the entire processed sequence to predict the next token.
The standard Transformer architecture manipulates only the representations of the elements of the input sequence. There are also Transformer-based neural network architectures that incorporate external memory. The Extended Transformer Construction (ETC) \citep{Ainslie:2020} employs a global-local attention mechanism to handle long inputs. The model input is separated into two parts: \textit{long input}, which is a standard Transformer input sequence, and a small set of auxiliary tokens called \textit{global input}. Hidden representations of the global input tokens store summarized information about sets of long input tokens. Each part of the input is associated with its type of attention: full self-attention between global input tokens, full cross-attention between global and long inputs, and self-attention restricted to a fixed radius for long input tokens. 

BigBird \citep{Zaheer:2021} and Longformer \citep{Beltagy:2020} sparsify attention from a quadratic to linear dependencе on the sequence length by serving a number of pre-selected input tokens to store global representations. Global tokens are allowed to attend the entire sequence and, as a result, accumulate and redistribute global information. In addition, BigBird can use extra tokens to preserve contextual information.

An extension of the input sequence with extra memory-dedicated tokens is implemented in MemTransformer, MemCtrl Transformer, and MemBottleneck Transformer \citep{MemTransformer:2020}. In MemTransformer, specially reserved memory tokens are concatenated with the encoder input sequence to form the Transformer input. The model uses full self-attention over the memory-augmented input sequence and processes inputs in a standard way. MemCtrl Transformer uses the same memory-augmented input sequence as MemTransformer and has a sub-network to control memory and original input sequence tokens separately. In MemBottleneck Transformer, full attention is allowed between the input and the memory only. So, to update the model, we firstly update memory representations, as presented in MemCtrl Transformer, and secondly update the sequence representation.

In our symbolic working memory model studied in this paper, input encoding follows the vanilla Transformer, but during the generation of an output sequence, a decoder decides whether to write the next token to the internal working memory or the output target prediction. Memory elements are represented with the same embeddings as for the tokens from the vocabulary to retain memory interpretability. Working memory tokens are processed as any other elements of the decoder input sequence, which allows the decoder to attend to both target and memory tokens.  

Working memory elements are tokens from the vocabulary, representing natural language words or subwords, which makes memory content more explainable. We can juxtapose the working memory content with golden target sentences and model target predictions and search for insights about the model decision-making process. We expect that after training, working memory should exhibit some properties that make it helpful in solving the target task. First, the working memory content should be related to the content of the target task. Second, the complexity of the memory content should correlate with the complexity of the task. In this paper, we study these properties of the working memory for Russian to English machine translation task on the corpora of various lexical and grammatical complexity.

We summarize the differences between the related work and the proposed method in Table~\ref{tab:related_comparison}.

\begin{table}[htpb]
\fontsize{7}{8}
\selectfont
\begin{center}
\begingroup
\setlength{\tabcolsep}{0.8pt}
\begin{tabular}{p{0.2\textwidth} | p{0.18\textwidth} | p{0.6\textwidth}}
\hline
Architecture & Memory form & Memory access \\
\hline
NTM \citep{Graves:2014}, GCL \citep{Meng:2018} & External memory is presented with uninterpretable memory matrix of fixed size. & NTM updates memory using the copy of memory from the previous time step. GCL does not use the previous memory states in the memory update procedure. Reading from the memory is processed by the recurrent network. GCL, unlike NTM, uses separated content and address components. \\ \hline
DNC \citep{Graves:2016}, Sparse DNC \citep{Rae:2016} & Memory is a matrix with dynamic allocation. & In DNC, writing to the memory and reading from it is done by differentiable attention mechanisms. In Sparse DNC, read and write operations are constrained to combine a constant number of non-zero memory entries. \\ \hline
Memory Networks \citep{Weston:2014}, End-to-End Memory Networks \citep{Sukhbaatar:2015}, HMN \citep{Chandar:2016} & Memory is an array storing vector representation of the input sequence. & Reading from memory in Memory Networks and End-to-End Memory Networks is done with recurrent attention. In HMN, memory is hierarchically structured to minimize computation when reading from memory. \\ \hline
D-NTM \citep{Gulcehre:2017} & Each memory matrix cell has content and trainable address vectors. & Memory addressing is location-based. Memory processing is done with an NTM-like controller network.\\ \hline
TARDIS \citep{Gulcehre:2017memory} & Memory matrix of a fixed size is controlled by an RNN similarly to NTMs and D-NTMs. & TARDIS uses discrete addressing when operating with memory. Writing information to the memory is done in the sequential order analogously to NTMs. When the memory is filled up, the access is based on tying the model write and read heads. \\ \hline
ETC \citep{Ainslie:2020}, BigBird \citep{Zaheer:2021}, Longformer \citep{Beltagy:2020} & Selected tokens from the encoder input sequence. & Writing to memory and reading from it is done with specific attention patterns. \\ \hline
MemTransformer, MemCtrl Transformer, MemBottleneck Transformer \citep{MemTransformer:2020} & A fixed number of special tokens is prepended to the encoder input sequence. & MemTransformer processes memory tokens as standard input tokens. MemCtrl Transformer reads from memory with the standard self-attention. Memory updates in MemCtrl Transformer and MemBottleneck Transformer are done with a special layer. To read from memory, the MemBottleneck Transformer input sequence attends only to memory. \\ \hline
Transformer with working memory in decoder (ours) & A fixed number of tokens from the\newline vocabulary is mixed with the target input. & The model-generated memory tokens are written to the decoder input sequence in positions corresponding to the their creation time steps. The memory reading mechanism is standard Transformer multi-head self-attention. \\ \hline

\end{tabular}
\endgroup
\end{center}
\caption{Comparison of the related MANN works and the Transformer with working memory.}
\label{tab:related_comparison} 
\end{table}

This work studies symbolic working memory in the Transformer decoder as a representation of pieces of information chosen by the neural model to make predictions and examines how this representation is related to the target task and how working memory improves the model performance. We believe that the addition of working memory into the Transformer decoder is the first attempt to store interpretable symbolic representations in external memory for a neural network model.

\section{Transformer with Working Memory in Decoder}

Transformer is a sequence-to-sequence encoder-decoder model. In a machine translation task, during inference, the Transformer input consists of two sequences of tokens: a sentence in the source language and its partial translation to the target language. The encoder part of the model processes the first sequence. Then, the Transformer decoder takes hidden representations from the encoder output and the piece of already generated translation to predict the next token.  

The standard Transformer decoder is stacked from $N$ identical layers. To process the $i$-th decoder layer, firstly, the normalized sum of target inputs $Y_{inp}$ and their masked multi-head attention scores $MHA(Q, K, V, mask)$ are calculated:    

\begin{equation}
A_{self,i} = LN(Y_{inp} + MHA(Y_{inp},Y_{inp},Y_{inp},look\_ahead\_mask)).
\end{equation}

Then the multi-head cross-attention between the sequence representation $A_{self,i}$ and the encoder output $E$ followed by normalization is done:

\begin{equation}
A_{cross,i} = LN(A_{self,i} + MHA(A_{self,i},E,E)).
\end{equation}

The aggregated representation $A_{cross,i}$ is updated with a position-wise feed-forward network $FFN(X)$, then a skip connection and normalization are used:

\begin{equation}
D_{out,i} = LN(A_{cross,i} + FFN(A_{cross,i})).
\end{equation}

To obtain logits, the $N$-th decoder layer outputs are sent to the final dense layer:

\begin{equation}
Y_{pred} = Linear(D_{out,N}).
\end{equation}

In working memory implementation  \citep{WorkingMemory:2022}, memory is represented by $M$ additional tokens in the decoder input. The Transformer decoder generates, stores, and retrieves $M$ working memory tokens in the same way it predicts the translation sequence. Memory tokens are placed in the decoder input sequence and processed by the model the same as standard Transformer decoder input tokens, so during decoding the sequence, the model has full access to the memory tokens generated so far.

To treat working memory tokens as a part of the Transformer decoder input sequence, we allow the positions of the memory tokens to be mixed with the positions of target predictions in the generated sequence. For every predicted token, the model also predicts whether the token will be stored in working memory or in the target sequence.

\begin{figure}[t]
  \centering
  \includegraphics[width=\textwidth]{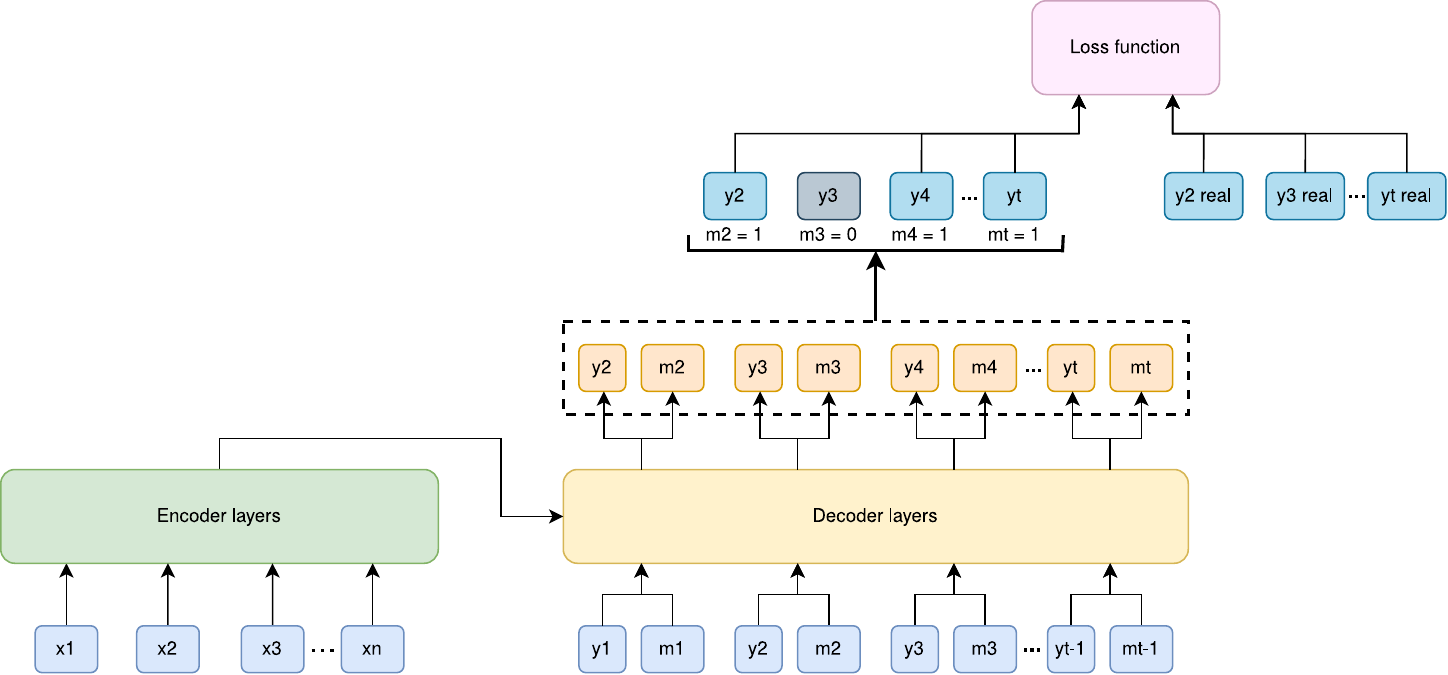}
  \caption{Transformer with the working memory-augmented decoder. The decoder inputs are the tokens generated to the moment $y_1,\dots,y_{t-1}$ and the corresponding memory flags $m_1,\dots,m_{t-1}.$ The memory flag is a binary value: $m_i=1$ means that $y_i$ is the target prediction token, and $m_j=0$ means $y_j$ is the working memory token. The final layer of the model has an expanded output size $= target\_vocabulary\_size+2$. The loss function takes into account the difference between the target sequence predictions and the real targets rather than the memory tokens.}
  \label{fig:model}
\end{figure}

The architecture is depicted in  Fig.~\ref{fig:model}. The model predictions are generated sequentially, one token at a time. When a newly generated token appears, the model decides if it is a memory token or the resulting translation token. Thus, the Transformer decoder input contains target sequence predictions alternating with memory tokens.

To allow the model to predict the token type and mark it with a dedicated flag value, we extend the dimensionality of the Transformer final layer up to $target\_vocabulary\_size + 2.$ Two additional units are used to predict the token type flag values. The embedding of this flag is added to the corresponding decoder input token embedding on the next decoding step to allow the model to differentiate the memory content from the target prediction values.

The procedure of training Transformer with working memory in decoder is described in Algorithm~\ref{alg:wm}. The model takes as an input the source sequence $X_{inp}$, the first token of the decoder input --- start-of-sequence token $Y_{inp} = (Y_1),$ start-of-sequence token type $T_{inp} = (T_1),$ ground truth target sequence $Y_{real},$ and the working memory size $mem\_size.$ The encoder transforms source sequence $X_{inp}$ into representation $E$. Then, given $E, Y_{inp}, T_{inp}$ and $Y_{real},$ the decoder generates the next token $Y_{pred}$ and its type $T_{pred}.$ According to the value of $T_{pred}$ and the number of memory tokens generated to the moment, the currently generated token is concatenated to $Y_{inp}$ with teacher forcing or just as is in the case of the token flagged as memory. The predicted token type value is appended to $T_{inp}.$
The sequence is generated token-by-token until the memory is full and the target prediction sequence matches the length of a real target.

\begin{algorithm}
\caption{Forward pass of Transformer with working memory in decoder}\label{alg:wm}
\fontsize{7}{8}
\selectfont
\begin{algorithmic}
\Require $X_{inp}, Y_{inp} = (Y_1), T_{inp} = (T_1), Y_{real}, mem\_size$
\State $E = Encoder(X_{inp})$
\State $i = 0, mem\_num = 0$
\While{$len(Y_{inp}) < len(Y_{real}) + mem\_size$}
    \State $(Y_{pred},T_{pred}) = Decoder(E, Y_{inp}, T_{inp})$
    \If{$T_{pred} == 0$} \Comment{$Y_{pred}$ token will be stored in memory}
        \If{$mem\_num < mem\_size$}
            \State $Y_{inp} = concat(Y_{inp}, Y_{pred})$
            \State $mem\_num = mem\_num + 1$
        \Else
            \State $T_{pred} = 1$
        \EndIf
    \EndIf
    \If{$T_{pred} == 1$} \Comment{teacher forcing target prediction token}
        \State $Y_{inp} = concat(Y_{inp}, Y_{real}[i + 1])$
        \State $i = i + 1$
    \EndIf
    \State $T_{inp} = concat(T_{inp}, T_{pred})$
\EndWhile
\end{algorithmic}
\end{algorithm}

For example, if the decoder input sequence is the following:
\begin{equation}
Y=[Y_1^{tar}, Y_2^{tar},Y_1^{mem},Y_3^{tar},Y_2^{mem},Y_{3}^{mem},Y_4^{tar}],
\end{equation}
 where $Y_i^{tar}$ are the target prediction tokens and $Y_j^{mem}$ are the tokens stored in the working memory, then the token type sequence for $Y$ will look as follows:
\begin{equation}
    T=[1,1,0,1,0,0,1].
\end{equation}
The token type vector helps locate the working memory elements in the predicted sequence. 

At the inference, there is no teacher forcing and every token and its type value generated by the model are stored in the decoder input sequence as is with the corresponding flag values.

In all experiments reported in this paper, memory size $M$ is set to 10. To calculate the loss function during training, we exclude the predicted sequence elements that belong to the working memory.
We use different decoding strategies for the target prediction tokens and the working memory content. To decode target predictions, we use the best path decoding and to obtain memory tokens, we apply nucleus sampling \citep{Holtzman:2020} with a sampling parameter $p_{nucleus}=0.9.$

\section{Datasets for Machine Translation Task}\label{experimental_setup}

This work aims to study the working memory content, find relations between model predictions and tokens stored in the memory, and explore how the complexity of an input text affects the working memory. For experiments, we used four datasets collected from different natural language domains. 

The first is the TED Ru-En machine translation dataset\footnote{\texttt{https://www.tensorflow.org/datasets/catalog/ted\_hrlr\_translate}} from the TED Talks Open Translation Project \citep{Ye:2018}. The TED dataset is a collection of transcripts of TED Talks, which are well-prepared speeches for a wide audience, so the sentences should be unambiguous, easy to understand, and grammatically correct at the same time. 

The second dataset consists of paired sentences from Russian Winograd Schema Challenge\footnote{\texttt{https://russiansuperglue.com/tasks/task\_info/RWSD}} (RWSD) and the original English Winograd Schema Challenge (WSC) dataset\footnote{\texttt{https://cs.nyu.edu/\textasciitilde davise/papers/WinogradSchemas/WS.html}}  \citep{Levesque:2012}. The Winograd schemas represent pairs of sentences that differ in only one or two words and contain an ambiguity that is resolved in opposite ways in two sentences and requires the use of world knowledge and reasoning for its resolution. Such ambiguity is also a challenge from the machine translation point of view because the translation system aims to keep the meaning of the sentence and make correct word choices to result in an accurate translation. Combining Russian and English versions of the Winograd schemas was possible because samples in Russian were collected by manually translating and adapting the original Winograd dataset for Russian. The translations were also human-assessed. 

The other two datasets are from the OPUS project \citep{Tiedemann:2012} and are sourced from TensorFlow Datasets\footnote{\texttt{https://www.tensorflow.org/datasets/catalog/opus}}.
Open Subtitles is a collection of translated movie subtitles\footnote{\texttt{https://www.opensubtitles.org}}. This dataset represents a collection of pairs of spoken language phrases and lines from movies. Such informal language includes colloquialisms, phrasal verbs, contractions that represent another level of translation complexity compared with the written language, as in WSC, or prepared speeches, as in TED.

IT documents is a collection of parallel corpora of localization files for GNOME, KDE4, and Ubuntu and documentation files for PHP and OpenOffice. The IT documents dataset consists of sentences written in technical language and contains field-specific terms and abbreviations. These dataset features make the task of machine translation of IT documents challenging. 

According to the described features of the data, in our further analysis, we consider the TED dataset as the least complex for a machine translation task, then Open Subtitles, WSC, and IT documents in the ascending order of translation difficulty.

The model pre-trained on TED was fine-tuned on WSC, Open Subtitles, and IT Documents for 30 epochs to test how memory content changes after domain adaptation. We inferred translations for sentences from the TED test set (5476 samples) and joined Winograd validation and test sets (95 samples) for the memory content study. There are only train sets available for Open Subtitles and IT documents, so we randomly selected 6000 samples that did not appear during fine-tuning from each dataset. The inference sets' sizes and sentence lengths are presented in Table~\ref{tab:data}.

\begin{table}[htpb]
\fontsize{7}{8}
\selectfont
\begin{center}
\begingroup
\setlength{\tabcolsep}{0.8pt}
\begin{tabular}{l c c c p{0.5mm} c c c p{0.5mm} c c c p{0.5mm} c c c}
\hline
\multirow{2}{*}{Value} & \multicolumn{3}{c}{IT En} & & \multicolumn{3}{c}{OpSub En} & & \multicolumn{3}{c}{TED En} & & \multicolumn{3}{c}{WSC En}\\
\cline{2-4} \cline{6-8} \cline{10-12} \cline{14-16}
& before & after & refs && before & after & refs && before & after & refs && before & after & refs \\ 
\hline\hline
Samples & \multicolumn{3}{c}{6000} && \multicolumn{3}{c}{6000} && \multicolumn{3}{c}{5476} && \multicolumn{3}{c}{95}\\ \hline
Min length
 & 2 & 2 & 12 && 2 & 4 & 12 && 2 & 2 & 2 && 14  & 14 & 16\\ \hline
Max length
 & 139 & 139 & 135 && 109 & 119 & 106 && 164 & 58 & 160 && 42  & 72 & 123\\ \hline
Average length
 & 24 & 23 & 31 && 17 & 19 & 18 && 20 & 20 & 23 && 30 & 34 & 45\\ \hline
\end{tabular}
\endgroup
\end{center}
\caption{Datasets for the working memory content study. For each dataset, we provide the size of the inference set, minimal, maximal, and average sample length in the tokens for the predicted translations before fine-tuning (column "before"), after fine-tuning (column "after"), and reference translations from the data (column "refs").}
\label{tab:data} 
\end{table}

Winograd English sentences are the longest on average, and the Winograd data has a fewer number of samples than the other three datasets. To equalize the range of possible translation lengths, we cut off the sentences from TED, Open Subtitles, and IT documents that are out of the bounds of the WSC predicted and reference translations length. We also drop the duplicating samples from the inference set. As a result, we analyze 4665 and 4875 samples from the IT dataset, 3874 and 4506 Open Subtitles' samples, 3477 and 3486 TED samples, and 95 and 95 WSC samples before and after fine-tuning, correspondingly.

The TED talks dataset used for the initial training is lowercase, and the samples from the IT documents, Open Subtitles, and Winograd Schema Challenge datasets have first letters of the first sentence word and proper nouns capitalized.

\section{Study of Memory Content}

The baseline model we study is Transformer with the working memory in the decoder that was pre-trained on the train set from the TED Ru-En dataset. For pre-training, we used a standard Transformer for the first five epochs and then added working memory to the decoder and continued pre-training for 15 epochs. We calculated BLEU 4 \citep{bleu:2002} and METEOR \citep{meteor:2007} scores averaged for three runs on the TED Ru-En validation set to evaluate the model.
The resulting model had $\text{BLEU}=21.30$ and $\text{METEOR}=48.81.$ This translation quality is slightly better than demonstrated by the standard Transformer model after 20 epochs $\text{BLEU}=21.16$ and $\text{METEOR}=40.93.$ The scores for all datasets before and after fine-tuning for the standard Transformer and the Transformer trained with working memory are presented in Table~\ref{tab:scores}.

\begin{table}[htpb]
\fontsize{7}{8}
\selectfont 
\begin{center}
\begingroup
\setlength{\tabcolsep}{0.8pt}
\begin{tabular}{l c c p{0.1mm} c c p{0.1mm} c c p{0.1mm} c c}
\hline
\multirow{2}{*}{Model} & \multicolumn{2}{c}{IT documents} & & \multicolumn{2}{c}{Open Subtitles} & & \multicolumn{2}{c}{TED} & & \multicolumn{2}{c}{WSC}\\
\cline{2-3} \cline{5-6} \cline{8-9} \cline{11-12}
& BLEU & METEOR && BLEU & METEOR && BLEU & METEOR && BLEU & METEOR \\ 
\hline\hline
Standard, pre-trained
 & 3.48 & 14.94 && 8.10 & 22.18 && 21.16 & 40.93 && 6.00 & 25.93\\ \hline
Standard, fine-tuned 
 & \textbf{7.85} & 17.20 && \textbf{11.71} & 22.37 && 22.10 & 41.89 && 6.76 & 23.54\\ \hline
WM, pre-trained
 & 3.42 & 15.26 && 8.20 & 22.27 && 21.30 & 48.81 && 6.58 & \textbf{27.78}\\ \hline
WM, fine-tuned
 & 7.83 & \textbf{17.25} && 11.55 & \textbf{22.58} && \textbf{22.18} & \textbf{49.38} && \textbf{7.50} & 27.09\\ \hline
WM, pre-trained,\\
masked mem
 &  &  &&  &  && 20.56 & 47.75 &&  & \\ \hline
\end{tabular}
\endgroup
\end{center}
\caption{Performance of the models with and without working memory in the decoder after pre-training and after fine-tuning for all datasets. The first two rows show the scores for the standard Transformer model. The third and the fourth rows correspond to the quality of predictions of the Transformer with working memory in the decoder. The last row shows the scores on TED for the model pre-trained with working memory for which the attention on memory tokens was disabled during inference. The best BLEU and METEOR scores for each dataset are highlighted in bold.}
\label{tab:scores} 
\end{table}

We also checked if the model trained with working memory uses it to improve the translation quality. We disabled attention on the working memory slots, so the memory was generated during inference, but the target sequence could not attend to it. The resulting metrics on the TED validation set were $\text{BLEU}=20.56$ and $\text{METEOR}=47.75$, which are lower on 0.74 BLEU and 1.06 METEOR points than the Transformer scores with working memory with a standard attention mechanism. This experiment shows the importance of working memory in the prediction process for better model performance.

Working memory in our experiments has a fixed size of 10 tokens, but the content of memory varies: it can be filled with a single repeating token or have several groups of identical tokens, or contain ten different tokens. Table~\ref{tab:preds_samples} shows the examples of sequences predicted by the model and the corresponding reference translations. 

\begin{table}[htpb]
\fontsize{7}{8}
\selectfont 
\begin{center}
\begingroup
\setlength{\tabcolsep}{0.8pt}
\begin{tabular}{p{0.08\textwidth} | p{0.24\textwidth} | p{0.36\textwidth} | p{0.28\textwidth}}
\hline
 Data & Source sentence (Ru) & Model prediction (En) & Reference translation (En)\\ \hline\hline
IT & \selectlanguage{russian}установить геометрию главного окна. & [start](scu)(install)(we )(because )\newline(add )(fi)(and )(so )(we)(set )we set up a geometry of the main window.[end] & sets the client geometry of the main widget.\\ \hline
IT & \selectlanguage{russian}Временная коллек- ция содержит файлы, которые вы на- значили на воспроизведение, но не хотите добавлять к какой-либо коллекции. & [start](the )(an )(but )(the )(the )\newline(we )(an )(going )(the )(it)the modern collections contains a file that you call it reproduction, but you don't want to add to any collection - any collection.[end] & The Temporary Collection is a collection that is used to hold songs you want to play but that you do n't want to add to any collection.\\ \hline
OpSub & \selectlanguage{russian}.. если они захотят проверить кабину. & [start](in )(and)(`` )(so )(.)(.)(now)\newline(.)(.)(`` )....if they want to check the cable test.[end] & ...if they want to check out the cabin.\\ \hline
OpSub & \selectlanguage{russian}Вы попросите кого-то из друзей сделать это для вас. & [start](ask )(people )(you )(just )\newline(ask )(take )(if )(you )(the )(so )you have to ask someone from friends to do it for you.......[end] & You're gonna get one of your judge friends to do it for you.\\ \hline
TED & \selectlanguage{russian}Нужно всё очень хорошо спланировать. & [start](it)(we )(we )(you )(we )(it )\newline(we )(we )(you )(it )we have to do all this very well.[end] & But it does require a lot of planning.\\ \hline
TED & \selectlanguage{russian}Знаю, что вас этому не учили, это не так легко заметить — но всё же попробуйте. & [start](now)(i )(i )(i )(and )(and )\newline(now )(i )(i )(i )i know you don't learn that, it's not so easy to notice — but still try.[end] &
I know that if you're not a trained brain expert, it's not that obvious, but just take a look, ok?
\\ \hline
WSC & 
\selectlanguage{russian}Мужчина не мог поднять своего сына, потому что он был слишком тяжел. & [start](the )(the )(the )(the )(the )\newline(the )(the )(the )(it )(the )the sales of the person couldn't rise up his son, because he was too bold.[end] & The man couldn't lift his son because he was so heavy.\\ \hline
WSC & Боб заплатил за обучение Чарльза в\newline университете. Но теперь Чарльз забыл об этом. Он не чувствует себя обязанным. & [start](however)(and )(sta)(the )\newline(they )(one )(the )(he )(and )\newline(higher )the prospect of learning to get closer to university, and the attorney now forgotten about that. the point is that it wasn't worth taking care of the same.[end] & Bob paid for Charlie's college education, but now Charlie acts as though it never happened. He is very ungrateful.
\\ \hline
\end{tabular}
\endgroup
\end{center}
\caption{Examples of the sequences of different lengths predicted by Transformer with working memory in the decoder and reference translations. The tokens stored in working memory are written in parentheses. [start] and [end] denote starting and ending tokens of the sequence, correspondingly. The remaining tokens represent the translation prediction.} 
\label{tab:preds_samples} 
\end{table}

\begin{figure*}[htbp]
    \centering
    \subfloat[Before fine-tuning]{
    \includegraphics[width=0.48\textwidth]{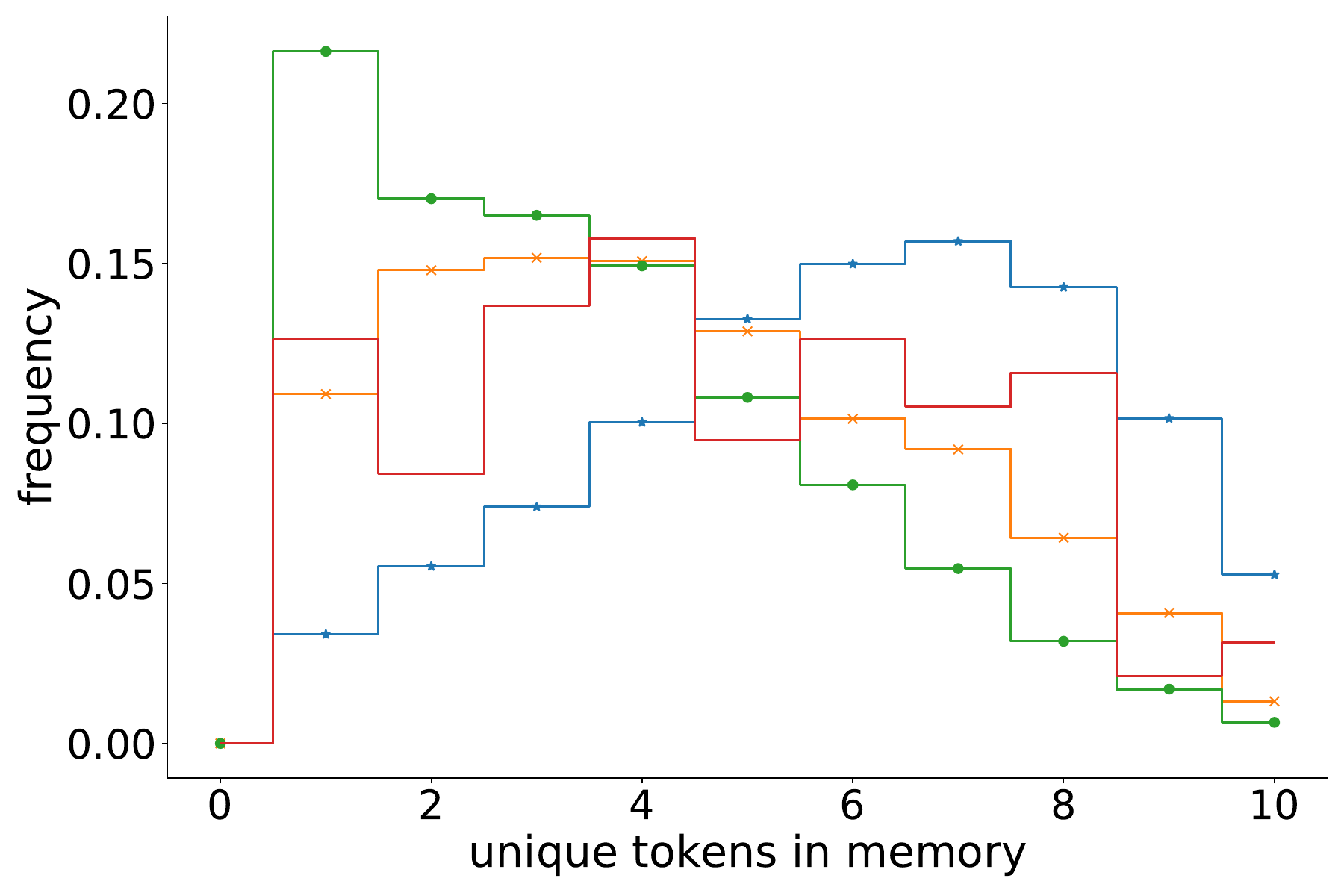}
    }
    \subfloat[After fine-tuning]{
    \includegraphics[width=0.48\textwidth]{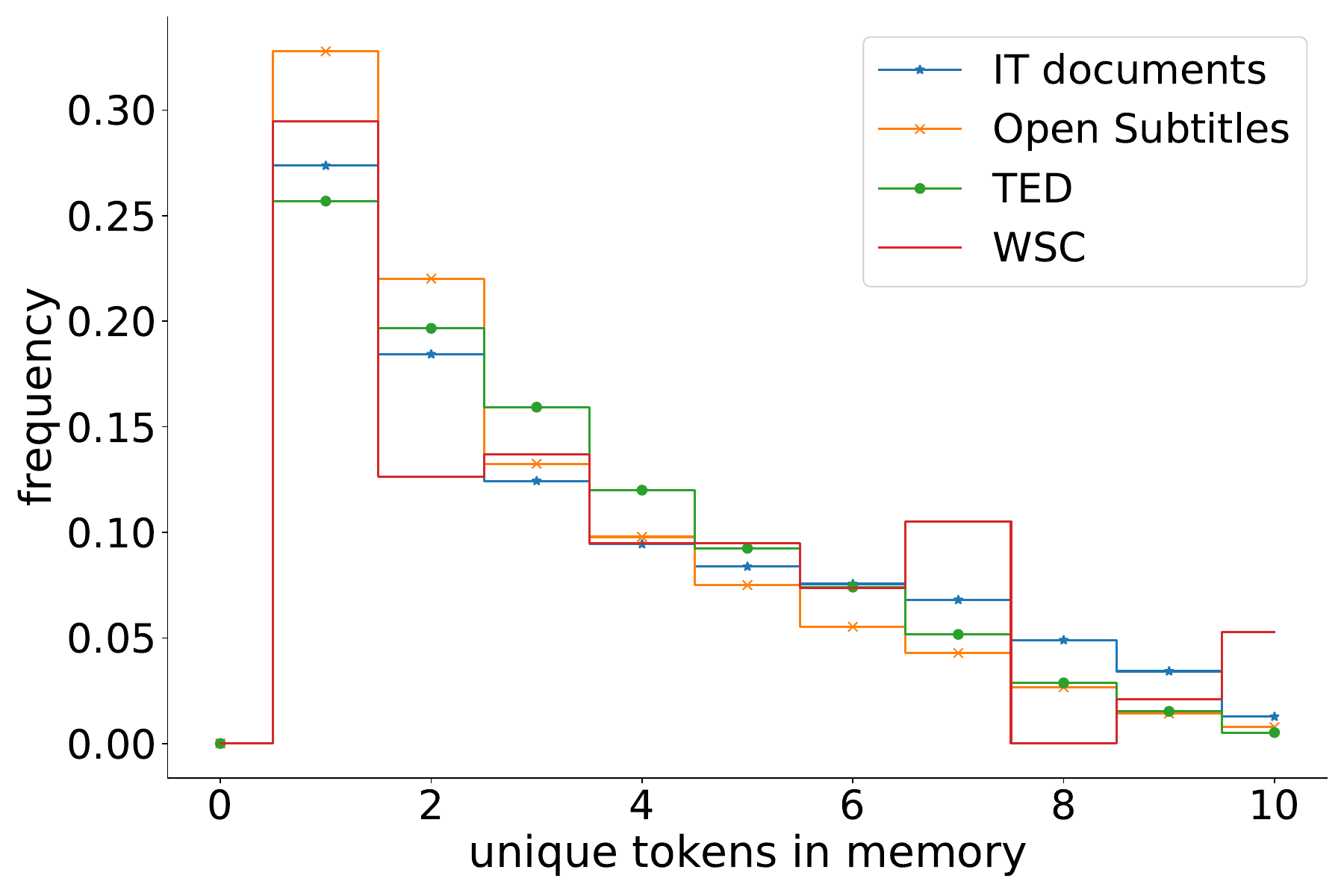}
    }
      \caption{Distributions of the number of unique tokens stored in working memory for the TED, WSC, IT documents, and Open Subtitles datasets. The histograms' legend before and after fine-tuning is presented in figure (b). Before fine-tuning, WSC, Open Subtitles, and IT documents memory diversity was larger than TED predictions' memory diversity. After fine-tuning, all datasets' working memory was mostly filled with a single repeating token. So, while processing the unseen data, the model exhibits higher variability of  the working memory content. More complex datasets demonstrate higher memory diversity.  After fine-tuning, the model was aligned with the data, and working memory had more repetitive tokens.}
  \label{fig:mem_distr}
\end{figure*}

It is natural to assume that translation of more difficult sentences should require more extensive utilization of working memory. To analyze the intensity of the working memory usage, we calculated the distributions of the number of unique working memory tokens. The histograms for all datasets before and after fine-tuning are presented in Fig.~\ref{fig:mem_distr}. 

From the histograms, we see that before fine-tuning, the memory content is more diverse for more complex sentences: the Open Subtitles and WSC working memory most frequently contains three and four different tokens, correspondingly, and the IT documents have seven different tokens in memory on average. On the other hand, the TED predictions' memory is most frequently filled with a single token. After fine-tuning, the model tends to store a single repeating token in memory most frequently for all datasets, sharpening the model attention on a specific term.

\begin{figure}[htbp]
    \centering
    \includegraphics[width=0.6\textwidth]{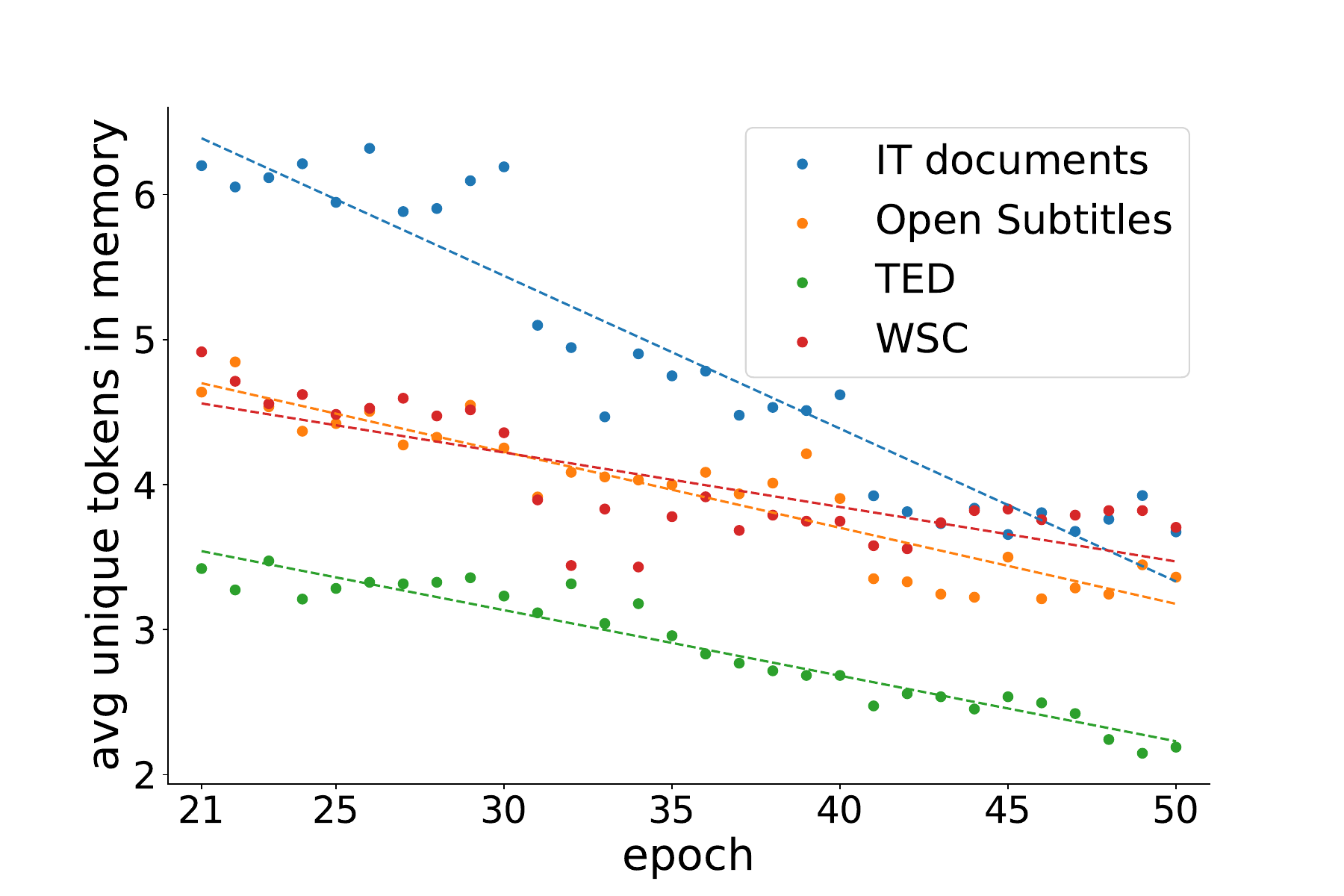}
    \caption{Average working memory diversity measured after each epoch of fine-tuning. The dashed lines are linear least squares fits. The plot confirms that during fine-tuning, the working memory content becomes more uniform. The minimal number of unique memory tokens is larger for more complex texts (IT docs and WSC) than for simpler texts (Open Subtitles and TED).}
    \label{fig:finetuning_avg_mem_diversity}
\end{figure}

We collected the model predictions after each epoch to explore the memory content behavior during fine-tuning. Figure~\ref{fig:finetuning_avg_mem_diversity} shows the average number of unique tokens stored in working memory for each fine-tuning experiment.

For all datasets, we see a diversity decrease in working memory during fine-tuning. The IT documents have higher overall memory diversity than the Winograd schemas and the Open Subtitles data, and the TED transcriptions have the lowest memory diversity. The IT documents contain the most field-specific texts, and the highest values of average memory diversity indicate that IT texts are very challenging to translate compared with the rest of the datasets used in our experiments.

Keywords are the most relevant and the most important words in a text. The keywords collection helps summarize the text and perceive the main topics discussed. So, we expected to find a higher number of keywords in memory for difficult IT and WSC datasets compared with Open Subtitles and TED.

\begin{figure*}[htbp]
    \centering
    \subfloat[Predictions' keywords in memory]{
    \includegraphics[width=0.48\textwidth]{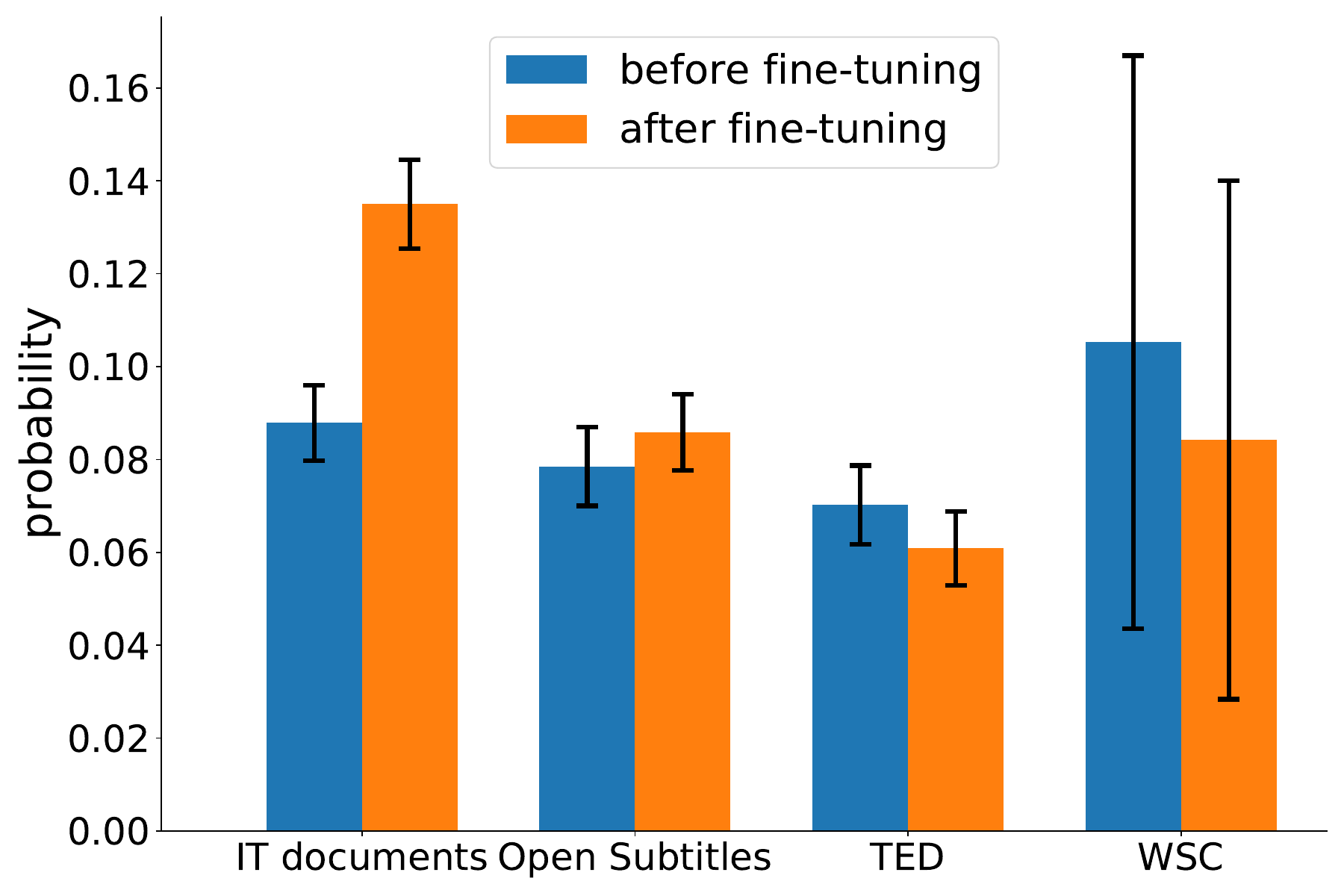}
    \label{fig:preds_kw}
    }
    \subfloat[Content words]{
    \includegraphics[width=0.48\textwidth]{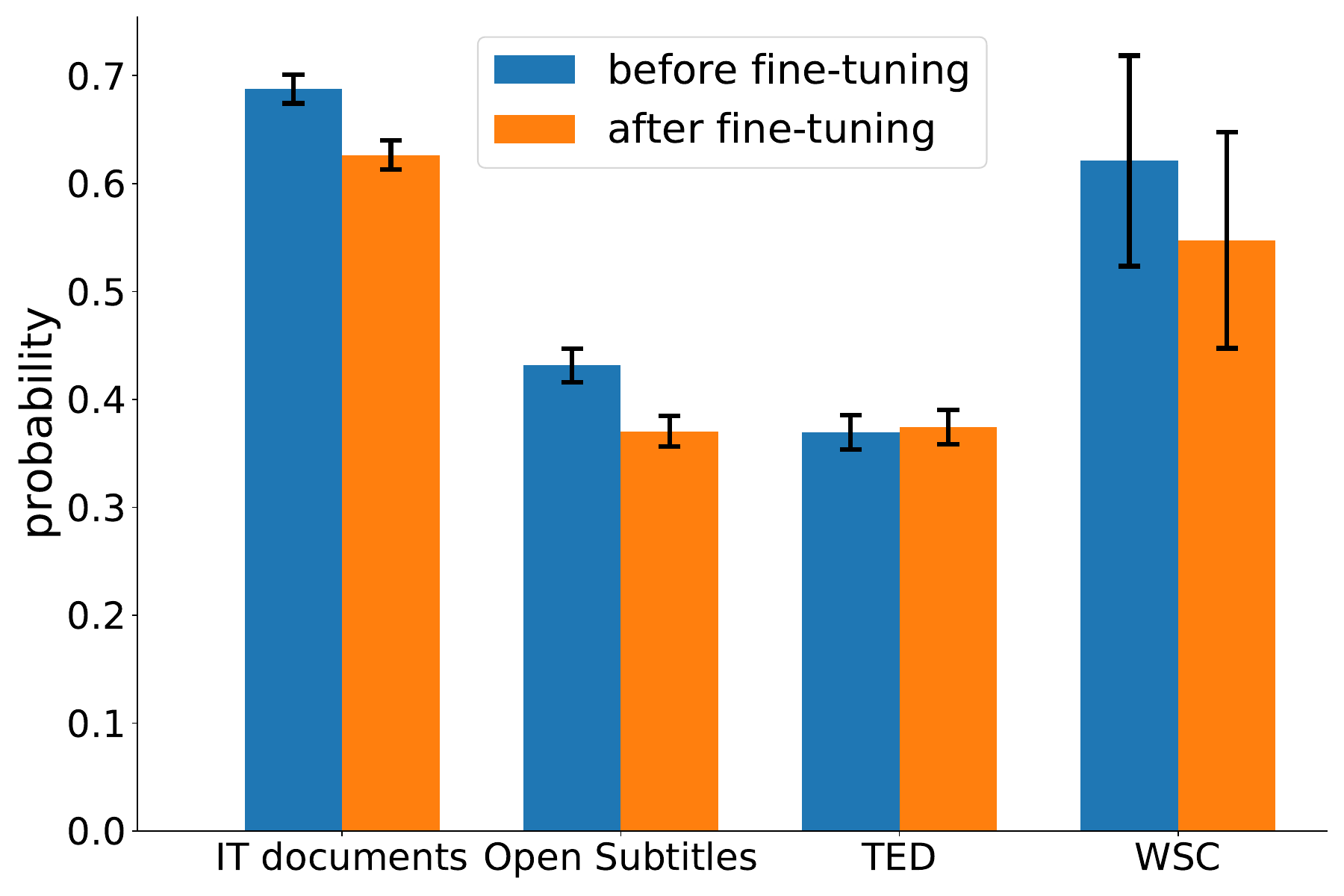}
    \label{fig:mem_keywords}
    }
    \caption{Probabilities (with confidence intervals) to find one or more (a) keywords extracted from the model predictions and (b) content words in working memory for all datasets. The keywords probability difference is significant for the IT documents-TED pair before and after fine-tuning $(p<0.01)$ and for IT documents-Open Subtitles pair after fine-tuning $(p<0.0001).$ Content words probabilities differ significantly for all complex-simple dataset pairs before and after fine-tuning $(p<0.01).$}
\label{fig:probas}
\end{figure*}

We extracted keywords from predicted translations and reference translations and calculated how many keywords are stored in working memory. The keyword extraction was made with the Rapid Automatic Keyword Extraction method (RAKE) \citep{rake:2010}. The resulting probabilities to find one or more keywords from the predicted sequence in memory are presented in Fig.~\ref{fig:preds_kw}. All datasets had at least one of the predicted sentence keywords in memory before and after fine-tuning. 
To assess the differences between keywords data, we used the Wilcoxon rank-sum test. We provide p-values for statistically significant differences. After fine-tuning, the IT documents had a significantly higher probability to store keywords in memory than before fine-tuning $(p<0.001).$ Overall, the more complex texts' memory stored predicted sequence keywords more often than the simpler texts' memory (the differences were statistically significant between the IT documents and TED datasets $(p<0.01)$ before and after fine-tuning and between IT documents and Open Subtitles after fine-tuning $(p<0.0001).$ Searching for reference sequence keywords in working memory, we found that similarly to the predictions' keywords, the IT documents probability to find keywords in memory significantly increased after fine-tuning $(p<0.05).$

Working memory could store words that possess semantic content. In linguistics, such terms are called \textit{content words}. Our hypothesis is that content words in memory represent key points of the text to be translated. We applied the keyword extraction method to memory to examine content words. The bar plot in Fig.~\ref{fig:mem_keywords} shows probabilities to find at least one content word in working memory. Similarly to the keywords probability analysis, the Wilcoxon rank-sum test was applied to compare content words data. All datasets before and after fine-tuning contain content words in working memory. In the IT documents' and WSC memory samples, content words appear significantly more frequently than in TED and Open Subtitles before and after fine-tuning ($p<0.01$ for pairs IT documents-Open Subtitles, IT documents-TED, WSC-Open Subtitles, WSC-TED. Each comparison was held before the fine-tuning procedure and after it).

\begin{figure*}[htbp]
    \centering
    \subfloat[Before fine-tuning]{
    \includegraphics[width=0.48\textwidth]{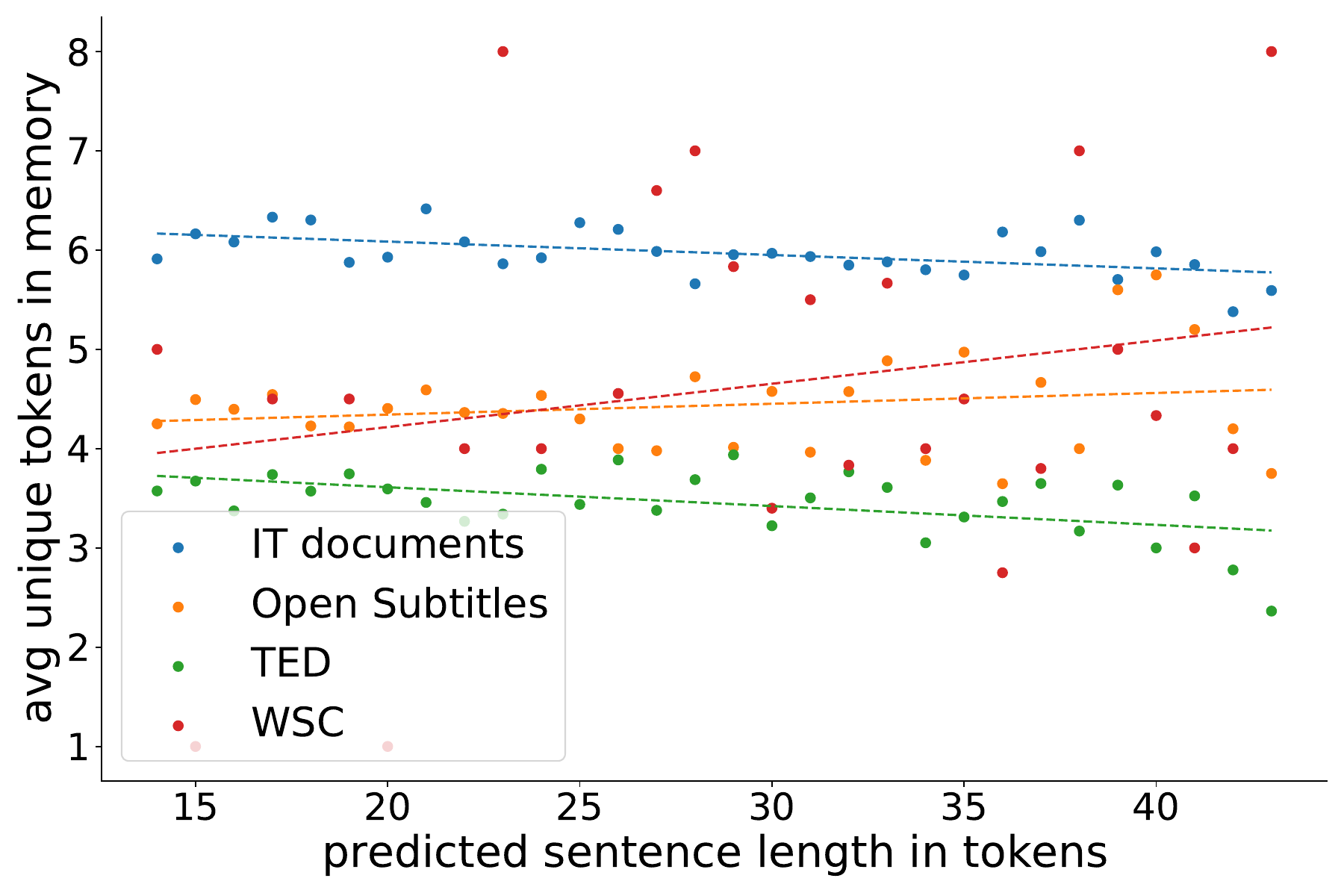}
    }
    \subfloat[After fine-tuning]{
    \includegraphics[width=0.48\textwidth]{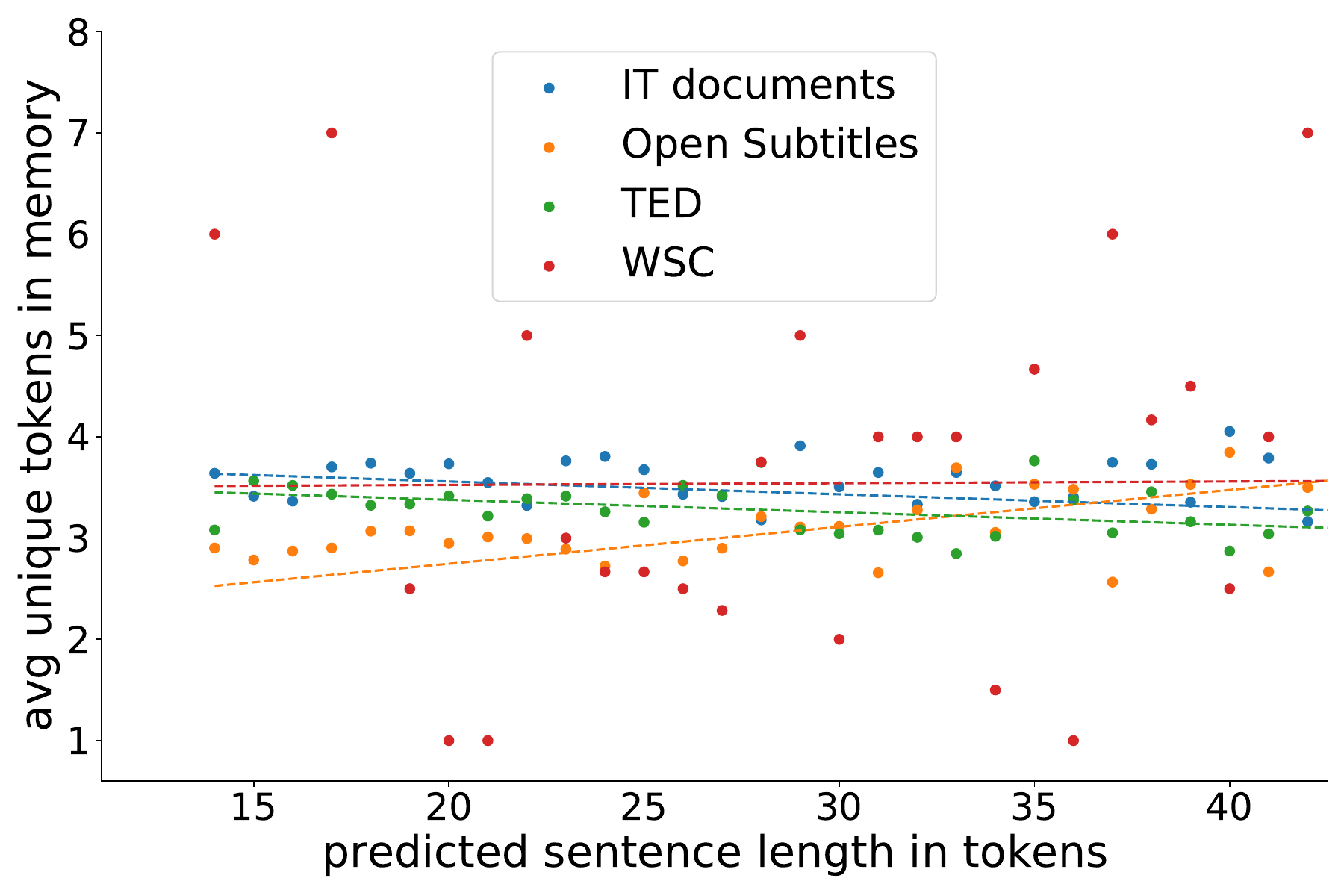}
    }
    \caption{Dependence of the average number of unique tokens in memory from the model predicted sequence length (with the dashed line showing a linear least squares fit). The average memory diversity does not significantly depend on the model prediction length both before and after fine-tuning.}
    \label{fig:avg_unique_mem_by_preds_len}
\end{figure*}

Longer sentences usually have more information, so they should be harder to translate. We checked how the average number of unique tokens in memory changes with the translation length (Fig.~\ref{fig:avg_unique_mem_by_preds_len}). We can see that the diversity of memory elements depends on the length of predicted translation length neither before nor after fine-tuning.

\begin{figure*}[t]
    \centering
    \subfloat[TED]{
    \includegraphics[width=0.22\textwidth]{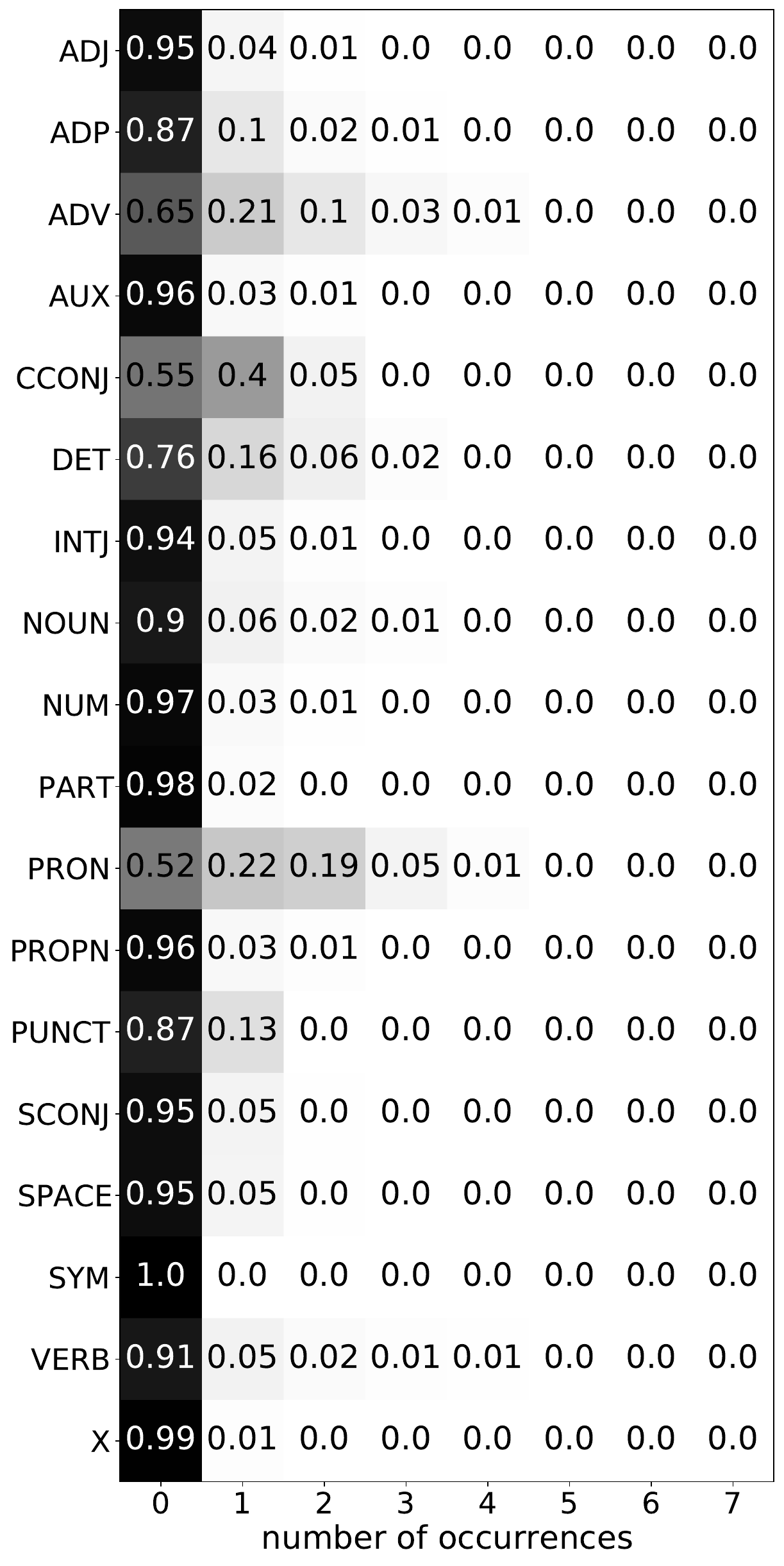}
    }
    \subfloat[WSC]{
    \includegraphics[width=0.22\textwidth]{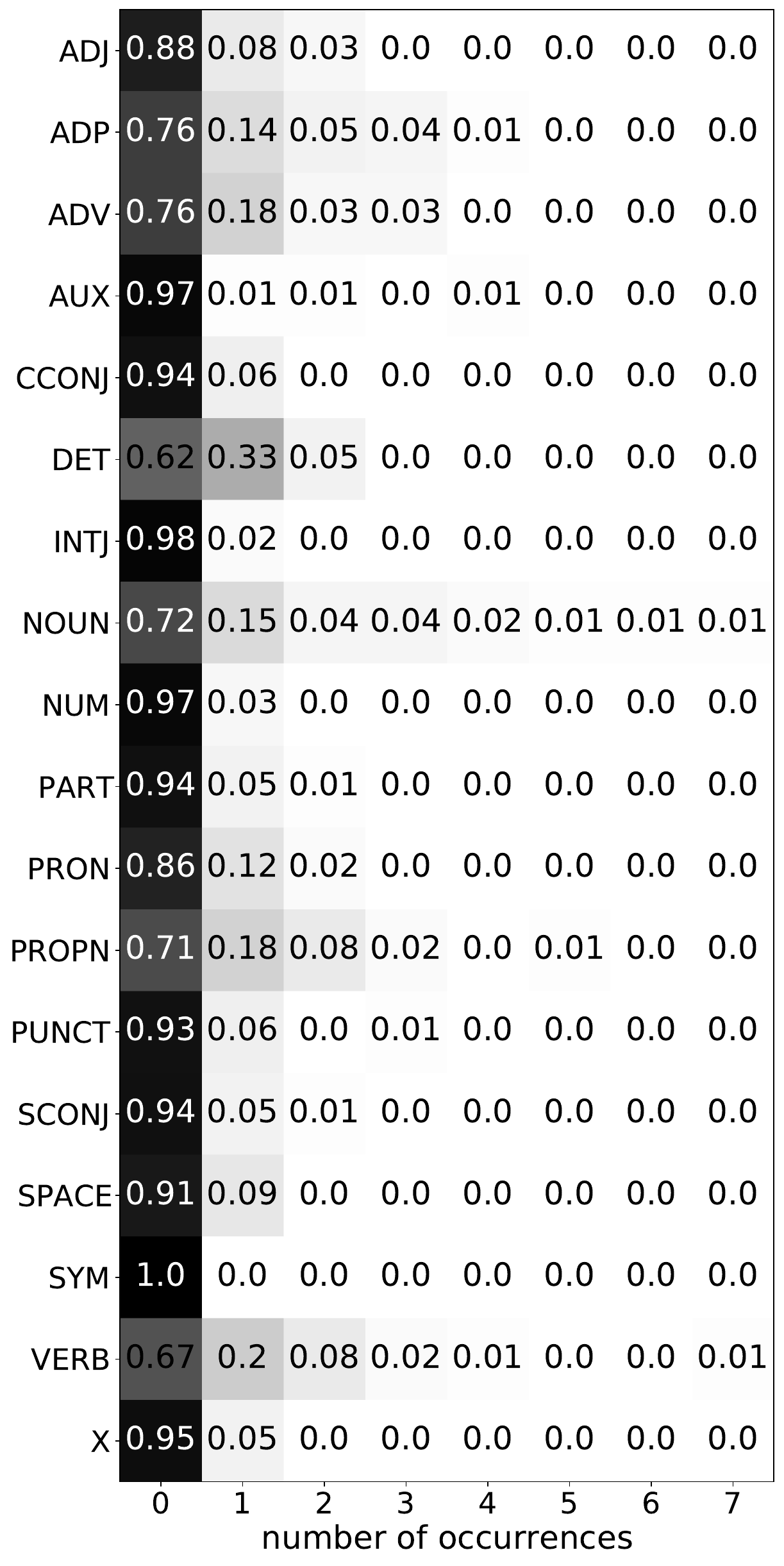}
    }
    \subfloat[Open Subtitles]{
    \includegraphics[width=0.22\textwidth]{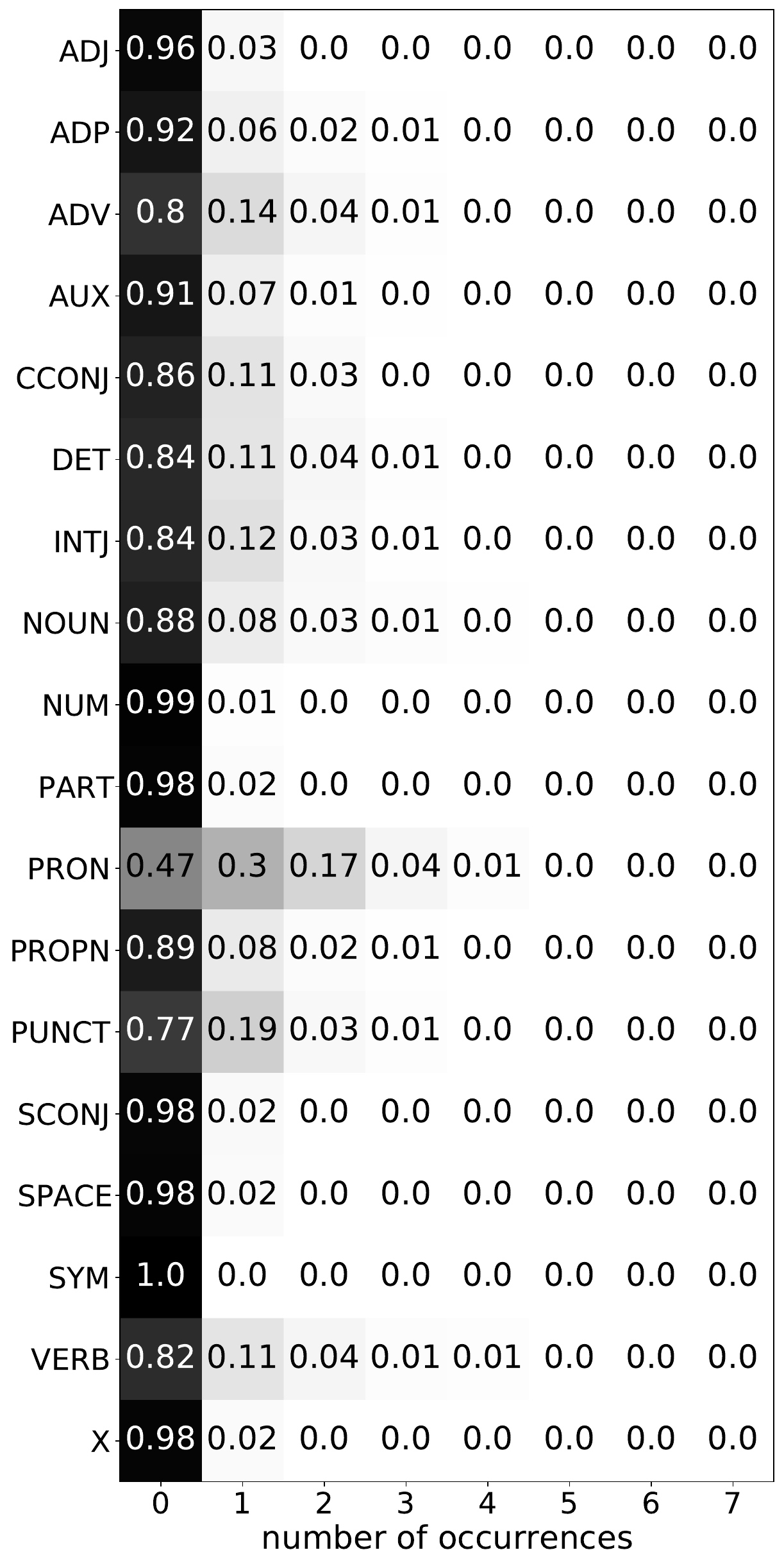}
    }
    \subfloat[IT documents]{
    \includegraphics[width=0.22\textwidth]{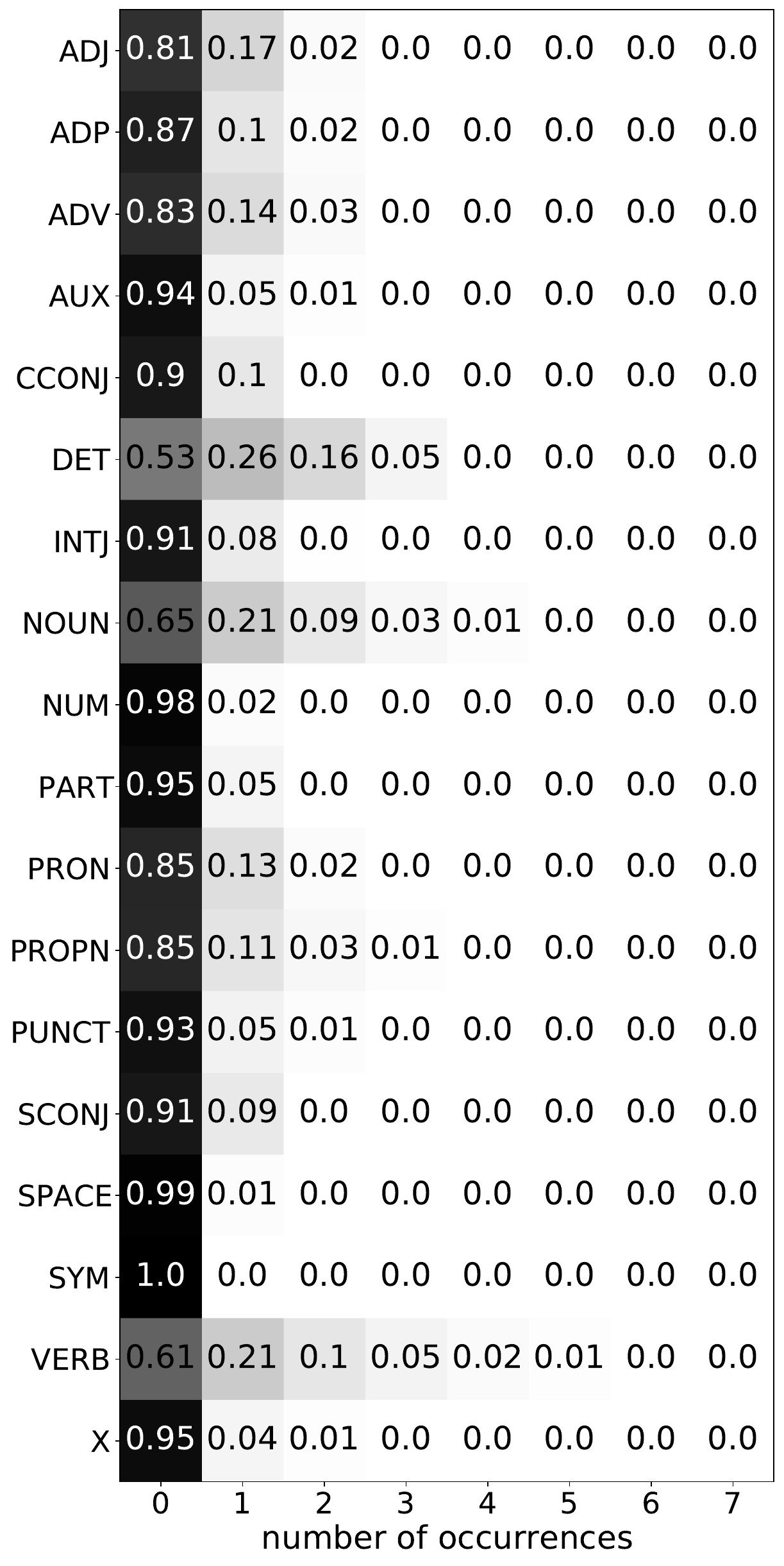}
    }
    \caption{Working memory parts of speech distributions for all datasets after fine-tuning. The distributions are showed up to seven occurrences because none of the examined parts of speech appeared more than seven times. After fine-tuning, the simpler texts memory (TED and Open Subtitles) contains coordinating conjunctions, pronouns, and punctuation marks. The WSC and IT documents' working memory most frequently comprises determiners, nouns, proper nouns, and verbs, similarly to the top parts of speech used in memory before fine-tuning.}
    \label{fig:parts_of_speech_after_ft}
\end{figure*}

Tokens written to working memory are the words or subwords of natural language. So likewise the unique tokens in memory, we can see what parts of speech are most likely written to working memory. We collected the distributions of parts of speech among unique tokens from memory after fine-tuning. Figure \ref{fig:parts_of_speech_after_ft} shows that memory tokens for the WSC and IT documents datasets significantly more often than TED and Open Subtitles store determiners, nouns, proper nouns, and verbs (the Wilcoxon rank-sum test, $p<0.05$ for all mentioned parts of speech for all complex-simple data pairs). TED and Open Subtitles also store significantly more coordinating conjunctions and pronouns compared to IT documents and WSC $(p<0.05).$ Punctuation marks occur significantly more often for TED and Open Subtitles compared to IT documents  $(p<0.0001)$ and for Open Subtitles compared to WSC $(p<0.0005).$

\section{Conclusion}

This work explored the features of the elements stored in the symbolic working memory of neural Transformer architecture. We compared the working memory content for a Russian to English machine translation task. We used IT documents, Open Subtitles TED Talks transcripts, and Winograd Schema Challenge datasets as examples of texts from different fields and different levels of translation complexity.

Firstly, we investigated if the information in memory is useful for solving a machine translation problem. We calculated how many unique tokens were stored in working memory most frequently and found that memory diversity is lower for simpler texts rather than for more complex ones. When the data sample appears in training for the first time, the maximum amount of information about the text is written into memory. The longer the model is being trained, the better it adjusts to the data, the less diverse the memory content becomes.

Secondly, during the working memory content analysis, we checked if the working memory content is relevant to the translated sentences. We calculated how often keywords extracted from translations occur in memory and found that at least one keyword occurs for all datasets. We also calculated the number of content words in working memory. Content words more often occur when translating more challenging texts containing ambiguous (WSC) or field-specific terms (IT documents). Finally, we found that the memory diversity decreases with the course of fine-tuning.

We examined parts of speech stored in memory: for more complex texts, determiners, nouns, proper nouns, and verbs occur more frequently than for less complex ones. This shows that memory is used to record information about grammar structures of more complex texts. 

\section*{Compliance with ethical standards}

Ethical approval: This article does not contain any studies involving human participants or animals performed by any of the authors.

\section*{Funding}

This work was supported by a grant for research centers in the field of artificial intelligence, provided by the Analytical Center for the Government of the Russian Federation under the subsidy agreement (agreement identifier 000000D730321P5Q0002) and the agreement with the Moscow Institute of\newline Physics and Technology dated November 1, 2021 No. 70-2021-00138.

\section*{Declaration of competing interest}

The authors declare that they have no known competing financial interests or personal relationships that could have appeared to influence the work reported in this paper.

\bibliography{main.bib}

\end{document}